\documentclass[12pt]{article}

\usepackage{scicite}
\usepackage[inline]{enumitem}
\usepackage{times}
\usepackage{graphicx}
\usepackage[font=footnotesize]{subcaption}
\usepackage[flushleft]{threeparttable}
\usepackage{hyperref}
\usepackage{xcolor}

\newenvironment{sciabstract}{%
\begin{quote} \bf}
{\end{quote}}

\title{A Metric for Characterizing \\the Arm Nonuse Workspace in Post-stroke Individuals \\ Using a Robot Arm} 

\author
{Nathaniel Dennler$^{1\ast}$,  Amelia Cain$^{2}$, Erica De Guzman$^{1}$, Claudia Chiu$^{1}$,\\ Carolee J. Winstein$^{2}$, Stefanos Nikolaidis$^{1}$, and Maja J. Matari\'c$^{1}$\\
\\
\normalsize{$^{1}$Department of Computer Science, University of Southern California,}\\
\normalsize{$^{2}$Department of Biokinesiology and Physical Therapy, University of Southern California,}\\
\\
\normalsize{$^\ast$To whom correspondence should be addressed; E-mail:  dennler@usc.edu.}
}

\date{}

\definecolor{beige}{RGB}{196, 196, 108}
\newcommand{\nathan}[1]{#1}

\begin{document} 


\baselineskip24pt


\maketitle

\begin{center}
\textcolor{red}{This manuscript was published in Science Robotics on November 15, 2023. This version has not undergone final editing. Please refer to the complete version of record at
\href{https://www.science.org/doi/10.1126/scirobotics.adf7723}{https://www.science.org/doi/10.1126/scirobotics.adf7723}. 
The manuscript may not be reproduced or used in any manner that does not fall within the fair use provisions of the Copyright Act without the prior, written permission of the American Association for the Advancement of Science (AAAS).}
\end{center}

\newpage

\begin{sciabstract}
  
  An over-reliance on the less-affected limb for functional tasks at the expense of the paretic limb and in spite of recovered capacity is an often-observed phenomenon in survivors of hemispheric stroke. The difference between capacity for use and actual spontaneous use is referred to as arm nonuse. Obtaining an ecologically valid evaluation of arm nonuse is challenging because it requires the observation of spontaneous arm choice for different tasks, which can easily be influenced by instructions, presumed expectations, and awareness that one is being tested. To better quantify arm nonuse, we developed the Bimanual Arm Reaching Test with a Robot (BARTR) for quantitatively assessing arm nonuse in chronic stroke survivors. The BARTR is an instrument that utilizes a robot arm as a means of remote and unbiased data collection of nuanced spatial data for clinical evaluations of arm nonuse.  This approach shows promise for determining the efficacy of interventions designed to reduce paretic arm nonuse and enhance functional recovery after stroke.  We show that the BARTR satisfies the criteria of an appropriate metric for neurorehabilitative contexts: it is valid, reliable, and simple to use. 
\end{sciabstract}

%
%

\section*{Introduction}
 
Stroke is a leading cause of serious long-term disability in the United States \cite{tsao2022heart}. Without sufficient rehabilitation efforts, functional decline will ensue, leading to increased difficulty in completing activities of daily living (ADLs), which contributes to decreased quality of life \cite{mayo2002activity, winstein2019dosage}. The goal of post-stroke neurorehabilitation is to restore functionality to the affected limb and enable stroke survivors to improve their quality of life. Several post-stroke rehabilitative interventions, such as task-oriented training \cite{rensink2009task}, biofeedback\cite{stanton2017biofeedback}, and constraint-induced movement therapy\cite{wolf2006effect}, have demonstrated substantial improvements along levels of the International Classification of Functioning, Disability and Health \cite{world2001ifc} including domains of body structure/function, activity limitations, and participation. 

Despite these functional improvements, a subset of stroke survivors may still experience a discrepancy between what they are able to do in tests where they are constrained to using their stroke-affected arm and what they spontaneously do in real world ADLs. This is of particular concern for individuals with hemiparetic stroke and other unilateral motor disorders, because the less-affected side can be used to compensate for movements of the impaired side and such compensation interferes with the “use it or lose it” foundational principle of neurorehabilitation. The nonuse phenomenon, the discrepancy between capacity and actual use \cite{taub1998constraint}, was first characterized in an article titled ``Stroke recovery: he can but does he?''~\cite{andrews1979sroke}. Nonuse has been shown to have a learned component \cite{buxbaum2020predictors}, and can thus be reduced through practice. This makes nonuse a key behavioral phenomenon to assess when evaluating patient recovery, one with high clinical and scientific significance.

In neurological rehabilitation contexts, outcome metrics must meet three criteria to be considered useful for evaluation: validity, reliability, and ease of use \cite{wade1992measurement}. However, the two currently widely-accepted instruments that provide metrics for nonuse--the Motor Activity Log (MAL)\cite{uswatte2006motor} and the Actual Amount of Use Test (AAUT)\cite{sterr2002neurobehavioral}--do not satisfy all three of those criteria. Although both tests have been found to be valid \cite{chen2012minimal}, they lack the other two desired qualities of neurorehabilitation assessment metrics: reliability and ease of use. \nathan{The MAL relies on a structured interview for user-reported arm use over the course of a specified duration; for example, one week or three days. Due to the difficulty associated with remembering and accurately describing one's arm use over the period of a week, this test is not simple for the participants to complete. The AAUT is a covert assessment that is valid only if the participant is unaware that the test is being conducted. Once the test is revealed, it becomes invalid for repeated use, making the scale unreliable.} Inspired by the current state of the field, this work introduces a metric for nonuse that meets all three criteria.

\begin{figure}[t!]
    \centering
    \includegraphics[width=.9\linewidth]{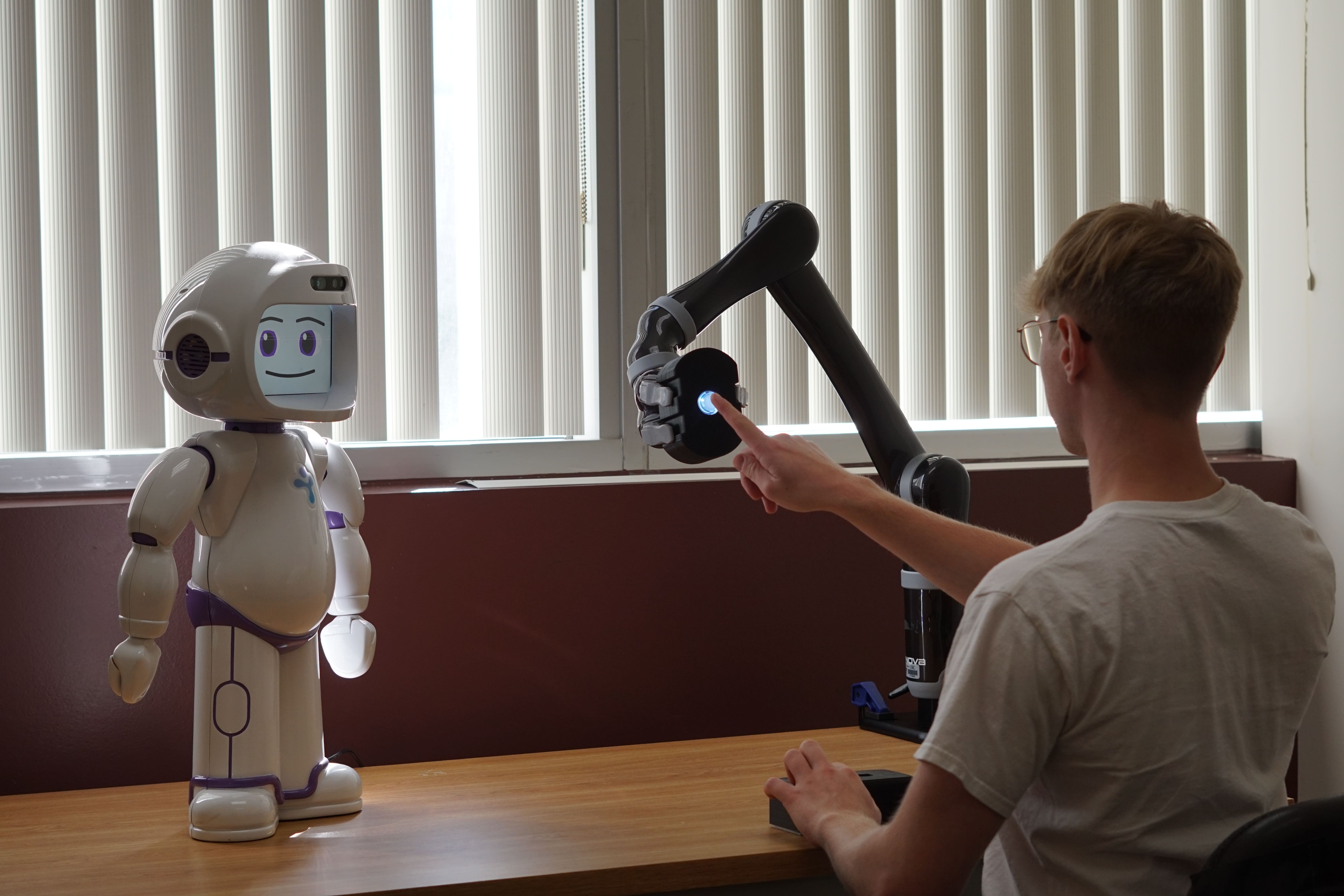}
    \caption{\textbf{Example reaching trial with the BARTR apparatus.} The participant places hands on the home position device. \nathan{The socially assistive robot (SAR, on the left) describes the mechanics of the BARTR, and the robot arm (on the right) moves the button to different target locations in front of the participant.} A reaching trial begins when the button lights up, and the SAR cues the participant to move.}
    \label{fig:study_setup}
\end{figure}

\nathan{Previous work demonstrated that the Bilateral Arm Reaching Test (BART) can be used to reliably quantify nonuse \cite{han2013quantifying}.} BART randomly lights up one of 100 equally-spaced points between 10cm and 30cm in front of the user, and the user is required to reach to the lit-up point within a time limit. In the first condition, the user is instructed to choose either hand to reach the point as quickly and as accurately as possible. Due to the imposed time limit, the user must make a fast and spontaneous hand choice, even if they know they are being tested. In the second condition, the user is constrained to only use their stroke-affected arm to reach for the point. The spontaneous performance in the first condition is compared to the functional performance in the second condition to assess the level of nonuse. This approach has been shown to be both reliable and valid; however, it only assesses patients on a single plane of motion. Reaching tasks required to accomplish ADLs involve three dimensional movements. In this study, we introduce a robot arm that enables a reaching task to quantify arm nonuse in three dimensions, allowing clinicians to tailor the rehabilitation process to specific patterns of nonuse as they occur in the user's real-world environment. 



We describe the modified Bimanual Arm Reaching Test with a Robot (BARTR), depicted in Figure \ref{fig:study_setup}. The testing apparatus consisted of a general-purpose robotic arm that queries points in front of the user, and a socially assistive robot (SAR) that supported the testing procedure by providing instruction and motivation. In a session of BARTR, the user completed two phases: a spontaneous phase and a constrained phase. Each phase can be completed in approximately 20 minutes. \nathan{We used identical instructions to the original validated BART \cite{han2013quantifying}. In the spontaneous phase, the user was instructed to use the hand that can reach the button as quickly and accurately as possible. These instructions ensured that participants acted spontaneously while being aware that they were being tested.} In the constrained phase, the user reached for the button with their stroke-affected hand. The nonuse metric, nuBARTR, was quantified from the reaching data collected from each session and repeated sessions that occurred at least four days apart, as in previous work \cite{han2013quantifying}.

To validate nuBARTR as a useful clinical metric, we developed the three following hypotheses based on the criteria for useful metrics in neurorehabilitation: First, nuBARTR is a valid metric, showing high correlation with the established metric for assessing nonuse, the AOU subscale of the AAUT. Second, nuBARTR is a reliable metric, posessing high test-retest reliability as evidenced by high absolute agreement across repeated sessions taken at least four days apart. Third, nuBARTR is a simple to use metric, achieving a score of 72.6 out of 100 or greater on the System Usability Scale, indicating above-average user experience as established in usability literature \cite{lewis2018system}.


We found that nuBARTR satisfies these three criteria for a useful neurorehabiliation metric: it had high validity, high test-retest reliability, and study participants found it easy to use. The system can be used to aid clinicians in the quantification and tracking of stroke survivor arm nonuse.

%
%

\section*{Results}

We performed a user study with neurotypical and post-stroke participants to evaluate the BARTR interaction. The nuBARTR was calculated from the BARTR interaction, and assessed for the properties of useful neurorehabilitation metrics.

\begin{table}[]
\centering
\begin{threeparttable}
    \caption{Demographic Information of the Post-Stroke Group}
    \begin{tabular}{lccc}
    \hline 
         & Median & Minimum & Maximum \\
         \hline
         FM-UE Motor Score (66 maximum) & 59.5 & 42 & 64 \\
         AAUT AOU Score (1 maximum) & .29 & .00 & .85 \\
         Age (years) & 55 & 32 & 85 \\
         \nathan{Time between sessions (days)} & \nathan{6.5} & \nathan{4} & \nathan{19}\\
         Gender & \multicolumn{3}{l}{8 Men, 6 Women}   \\
         Affected Side & \multicolumn{3}{l}{5 Left, 9 Right}   \\
         Ethnicity &  \multicolumn{3}{l}{4 Asian, 2 Black, 4 Hispanic,} \\
         &  \multicolumn{3}{l}{3 White, 1 Mixed-race} \\
         \hline
    \end{tabular}
    \begin{tablenotes}
      \small
      \item Abbreviations: FM-UE, Fugl-Meyer Upper Extremity; AAUT AOU, Actual Amount of Use Test -- Amount of Use
    \end{tablenotes}
    \label{tab:demographics}
    \end{threeparttable}
\end{table}

\subsection*{Participant Demographics and Stroke Characteristics}
Participants with chronic stroke were recruited from the Los Angeles, California, USA area to take part in this study. Participants were recruited through the IRB-approved Registry for Aging and Rehabilitation Evaluation database of the Motor Behavior and Neurorehabilitation Laboratory at the University of Southern California (USC). All participants were right-hand dominant prior to their stroke. In total, 17 post-stroke participants were recruited. Two participants did not meet the study criteria after screening and one participant was excluded from analysis due to difficulties in completing the task. Of the fourteen eligible participants, twelve completed all three sessions of the BARTR, and two were only able to complete two sessions due to scheduling constraints. One eligible participant did not receive AAUT scores due to technical problems in recording the exam. The average age of post-stroke participants was 57 $\pm$ 11 years. Age and other participant demographic information is summarized in Table \ref{tab:demographics}.

\begin{table}[]
\centering
    \caption{Demographic Information of the Neurotypical Group}
    \begin{tabular}{lccc}
    \hline 
         & Median & Minimum & Maximum \\
         \hline
         Age (years) & 69.5 & 45 & 82 \\
         Gender & \multicolumn{3}{l}{5 Men, 5 Women}   \\
         Ethnicity &  \multicolumn{3}{l}{2 Asian, 2 Black, 6 White} \\
         \hline
    \end{tabular}
    \label{tab:neurotypical_demographics}
\end{table}

We also recruited 10 neurotypical adults to establish a normative value for performance. All neurotypical participants were right-hand dominant, and their demographic information is summarized in Table \ref{tab:neurotypical_demographics}. The average age of neurotypical participants was 67 $\pm$ 10 years. 



\begin{figure}[t!]
  \centering
\includegraphics[width=\linewidth]{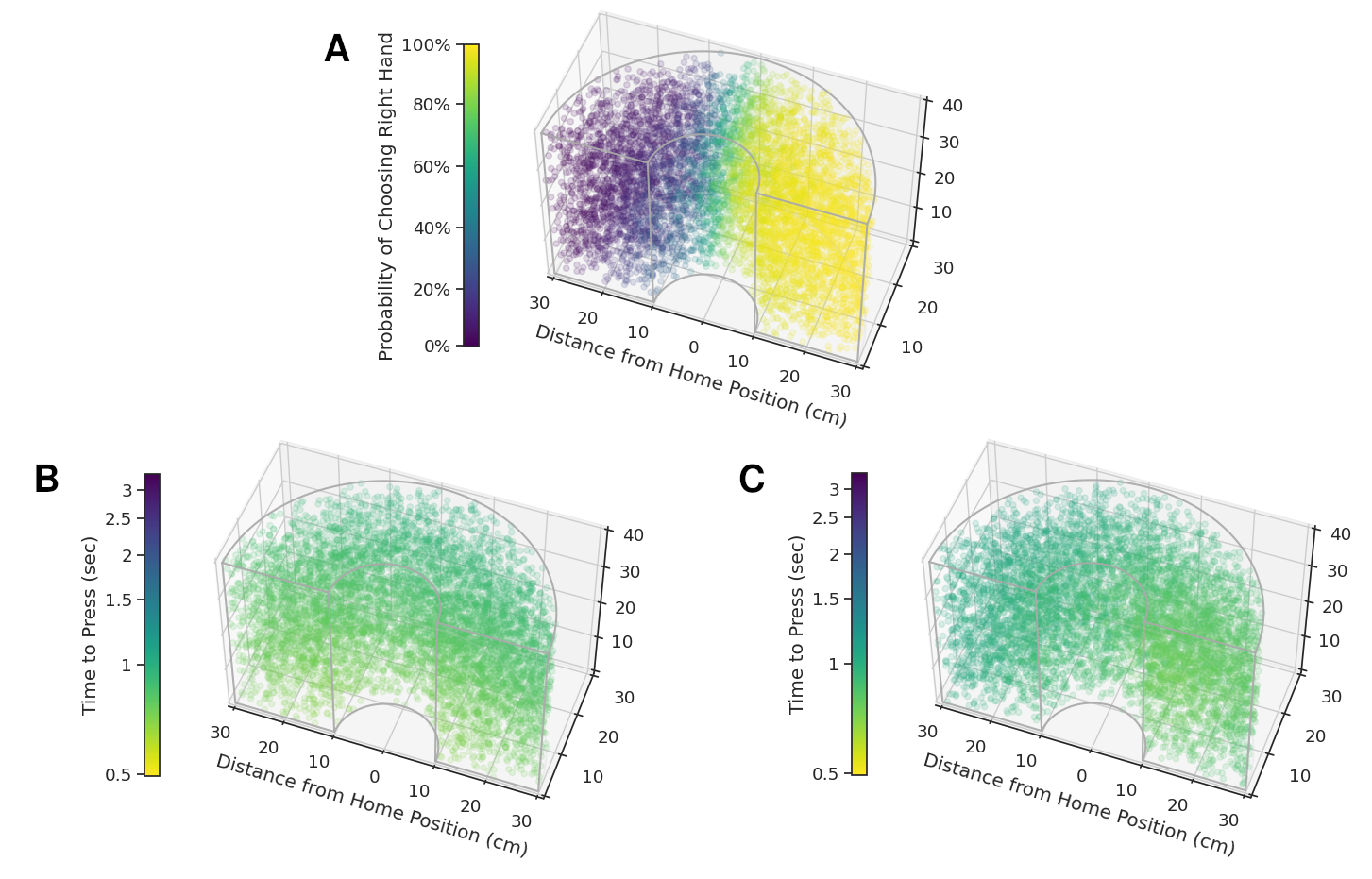}
    

    
  \caption{\textbf{Normative data collected from neurotypical participants.} The normative data consists of hand choice in the spontaneous condition (\textbf{A}) and average time for participants to reach with their left hand (\textbf{B}) and their right hand (\textbf{C}). Lighter colors indicate high probability of participants choosing their right hand (\textbf{A}) or faster times to reach (\textbf{B}, \textbf{C})}
   \label{fig:neurotypical_workspace}

\end{figure}

\subsection*{BARTR Use Characteristics}
We performed multiple analyses to assess the validity of the BARTR interaction as a nonuse metric.

\subsubsection*{Arm Use Characteristics}

To establish a normative baseline for comparisons, we examined the reaching data (time to reach and hand choice) from the 10 neurotypical participants. \nathan{We found that, for the neurotypical group, there were no significant differences in average time from leaving the starting position to pressing the button across participant age ($r^2=.06, p=.498$) or gender ($r^2=.03, p=.511$), as evidenced by linear regressions.} Similarly, for the neurotypical group, hand choice in the spontaneous condition had no significant differences due to participant age ($r^2=.004, p=.861$) or gender ($r^2=.015, p=.739$). Given the similarities across this group in performance on the BARTR task, we developed a single model of normative use based on an aggregate of the neurotypical participants' data. 

A visualization of the neurotypical group's interaction metrics is shown in Figure \ref{fig:neurotypical_workspace}. On average, there was a handedness bias, where the right side was used to press the button in 60\% of the workspace across participants, whereas the left side was used to press the button in 40\% of the workspace, identically to the 60-40 handedness bias reported in the planar BART \cite{han2013quantifying}. In general, the time to reach the targets was relatively consistent for both hands, with farther points taking slightly more time, as expected.

\begin{figure}[t!]
  \centering

    

    \includegraphics[width=\linewidth]{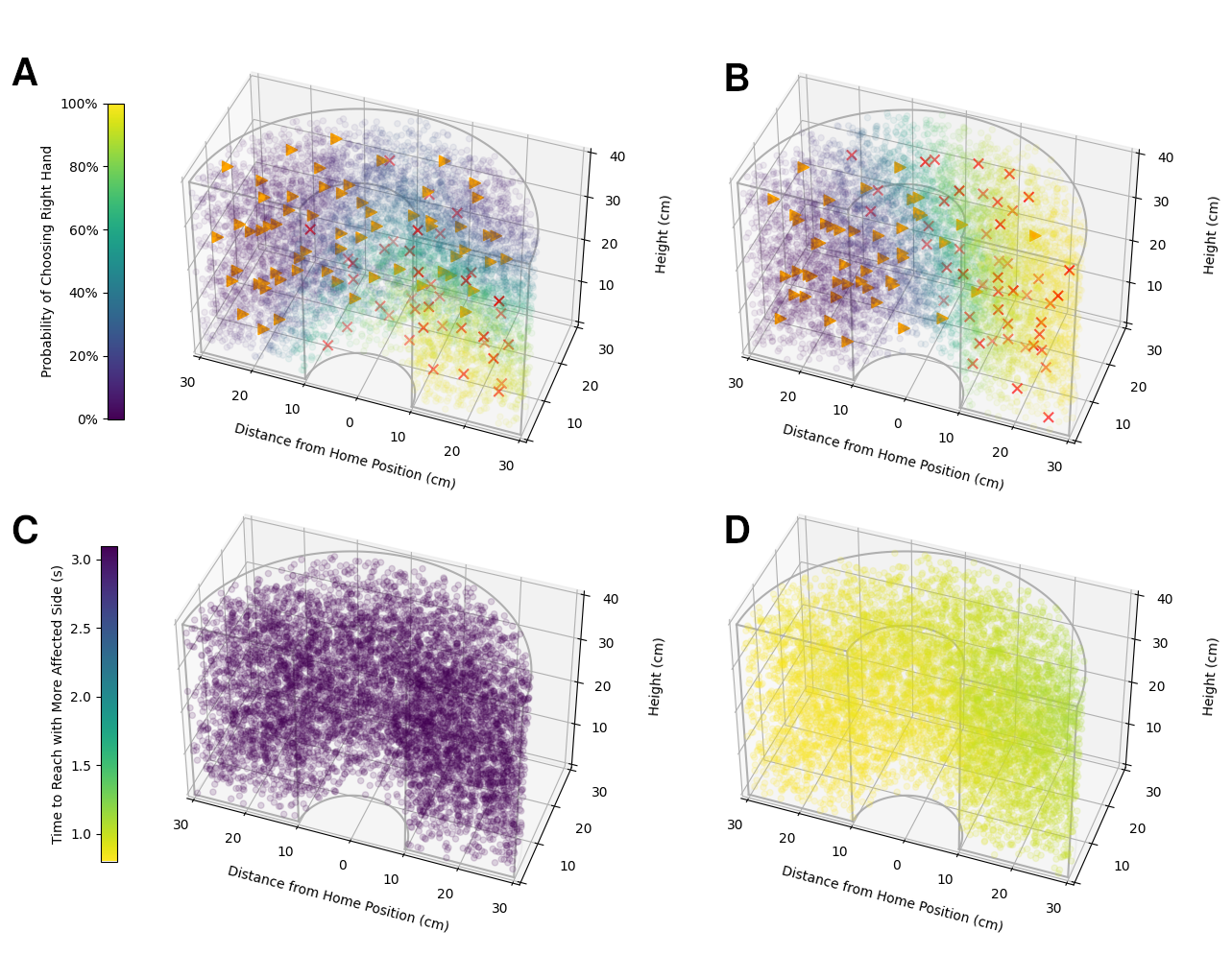}
    
  \caption{\textbf{Comparisons of data collected from two participants.} P23 was right-dominant affected, and showed lower right arm use (\textbf{A}) as well as longer reaching times (\textbf{C}). P31 was left non-dominant affected and showed more balanced right arm use (\textbf{B}) and faster reaching times (\textbf{D}). Raw data is shown for arm choice data, with a red `x' denoting right hand reaches, and an orange triangle denoting left hand reaches (\textbf{A},\textbf{B}). }
   \label{fig:specific_participants}

\end{figure}

In chronic stroke survivors, we observed high variability in hand choice and in the time to reach targets in the workspace. For illustrative purposes, in Figure~\ref{fig:specific_participants} we show these two interaction metrics modeled by Gaussian processes for two participants: one who was right-dominant affected and had high nonuse (P23), and one who was left-non-dominant affected and had low nonuse (P31). 

These data plots highlight the importance of including three-dimensional movement in the evaluation of nonuse. For example, P23 exhibits lower use of the right hand (63\% left handed, 37\% right handed), specifically in areas that appear higher on the right side, but maintains a high probability of using the affected arm for lower areas on the same side. P31 exhibits more symmetric use (37\% left handed, 63\% right handed), but also uses the less-affected side slightly more often for points that are higher up and closer to the mid-line.


\begin{table}[!ht]
    \centering
    \hspace*{-3em}
    \begin{threeparttable}
    \caption{Modeling Results for the Three Interaction Metrics} \label{table:classification_results}
    \begin{tabular}{lccc|ccc|ccc}
    \hline
         &\multicolumn{3}{c}{Side Classifier} & \multicolumn{3}{c}{Success Classifier} & \multicolumn{3}{c}{Time Regressor} \\ 
        Kernel & ACC$\uparrow$ & NLL$\downarrow$ & NLML$\downarrow$  & ACC$\uparrow$ & NLL$\downarrow$  & NLML$\downarrow$  & MSE$\downarrow$  & ME$\downarrow$  & NLML$\downarrow$  \\  \hline
        $  k_l + N_1$ & .838 & \textbf{.330} & 31.100 & .909 & .193 & 29.829 & .545 & 1.306 & 72.005 \\ 
        $  k_l + N_2$ & .838 & .332 & 31.018 & .909 & .195 & 29.539 & .545 & 1.307 & 69.675 \\ 
        $  k_l + N_3$ & .837 & \textbf{.330 }& 31.100 & .909 & .193 & 29.829 & .545 & 1.306 & 72.005 \\ 
        $  k_{rbf} + N_1$ & .838 & .342 & 31.664 & \textbf{.912} & .193 & 29.039 & .542 & 1.314 & 66.644 \\ 
        $  k_{rbf} + N_2$ & .836 & .342 & 31.637 & \textbf{.912} & .196 & 29.003 & .546 & 1.313 & 64.941 \\ 
        $  k_{rbf} + N_3$ & .836 & .342 & 31.660 & \textbf{.912} & .193 & 29.039 & .541 & 1.317 & 66.417 \\
        $  k_l +  k_{rbf} + N_1$ & .837 & .331 & \textbf{30.814} & .910 & .193 & 28.933 & \textbf{.540} & 1.308 & 66.762 \\ 
        $  k_l +  k_{rbf} + N_2$ & .838 & .334 & 30.859 & .910 & .196 & 28.899 & \textbf{.540} & 1.306 & 65.210 \\ 
        $  k_l +  k_{rbf} + N_3$ & \textbf{.839} & \textbf{.330} & 30.865 & .910 & .193 & 28.933 & .541 & 1.307 & 66.762 \\ 
        $  k_{comb.} + N_1$ & .834 & .345 & 31.209 & .910 & .193 & 29.071 & .542 & 1.305 & 66.462 \\ 
        $  k_{comb.} + N_2$ & .836 & .346 & 31.274 & .911 & .195 & 29.017 & .541 & 1.304 & 64.863 \\ 
        $  k_{comb.} + N_3$ & .836 & .345 & 31.268 & .911 & \textbf{.192} & 29.046 & .542 & \textbf{1.303} & 66.462 \\ 
        $  k_l +  k_{rbf} +  k_{comb.} + N_1$ & .837 & .342 & 30.870 & .910 & \textbf{.192} & 28.942 & .542 & 1.310 & 65.978 \\ 
        $  k_l +  k_{rbf} +  k_{comb.} + N_2$ & .836 & .341 & 30.983 & .910 & .194 & \textbf{28.858} & .541 & 1.307 & \textbf{64.718} \\ 
        $  k_l +  k_{rbf} +  k_{comb.} + N_3$ & .836 & .343 & 30.983 & .910 & .193 & 28.927 & .541 & 1.312 & 66.110 \\ 
        
         \hline
        Non-GP Method & ~ & ~ & ~ & ~ & ~ & ~ & ~ & ~ & ~ \\ \hline
        AdaBoost & .809 & .570 & - & .889 & .264 & - & .586 & 1.399 & - \\ 
        k-NN & .832 & 1.131 & - & .899 & .731 & - & .614 & 1.457 & - \\ 
        MLP & \textbf{.842} & \textbf{.331} & - & .908 & .195 & - & \textbf{.568} & \textbf{1.368} & - \\ 
        Random Forest & .834 & .437 & - & .899 & .327 & - & .617 & 1.476 & - \\ 
        Linear SVM & .834 & .332 & - & .905 & \textbf{.192} & - & .606 & 1.453 & - \\ 
        RBF SVM & \textbf{.842} & .339 & - & \textbf{.914} & .193 & - & .623 & 1.407 & - \\ \hline
    \end{tabular}
    \begin{tablenotes}
      \small
      \item Arrows indicate direction of better fits. Bolded values represent best values for each column. Abbreviations: ACC, accuracy; NLL, negative log likelihood; NLML, negative log marginal likelihood; MSE, mean squared error; ME, maximum error; $k_l$, linear kernel; $k_{rbf}$, radial basis function kernel; $k_{comb.}$, $k_l*k_{rbf}$; $N_1$, constant noise; $N_2$ constant noise + linear noise; $N_3$, constant noise + radial basis function noise; k-NN, k-nearest neighbors; MLP, multi-layer perceptron; SVM, support vector machine.
    \end{tablenotes}
    \end{threeparttable}
\end{table}

\subsubsection*{Selecting Kernels for Gaussian Process Modeling}

Three quantities were modeled through Gaussian processes to calculate arm nonuse: reaching success in the constrained phase of the BARTR, arm choice in the spontaneous phase of the BARTR, and reaching time for the affected arm across both phases of the BARTR. Success and arm choice are classification problems that leverage the Laplace approximation to model a non-Gaussian posterior with a Gaussian process, as is standard for classification~\cite{rasmussen2006gaussian}.  Reaching time is modeled directly as a regression problem. The results across several kernel choices for the Gaussian processes are shown in Table \ref{table:classification_results}.

In the context of the reaching task, both the distance the hand travels and the spatial location of the target are important for predicting time to reach and the selected reaching arm \cite{roby2003motor,han2013quantifying}. The kernels we tested were composed of two key components: the linear kernel, and the radial basis function kernel. The linear kernel indicates that points of similar distances from the origin will have similar values, and is defined as:
\begin{equation}
    k_{l}(x,x') = \sigma_0^2 + x\cdot x'
\end{equation}
where $\sigma_0$ is a hyperparameter learned from the data.

The radial basis function kernel indicates that points near each other in space will have similar values, and is defined as:
\begin{equation}
    k_{rbf}(x,x') = exp(-\frac{d(x,x')^2}{2\ell^2})
\end{equation}
where $\ell$ is the length-scale hyperparameter learned from the data.

The kernels were combined with addition and multiplication to represent different relationships between distance and locality, in accordance with recommendations from Duvenaud et al. \cite{duvenaud2013structure}. In total, 15 kernels were tested that combined 5 kernels that encapsulated the signal: $k_l, k_{rbf}$,$ k_l+k_{rbf}$,$ k_l*k_{rbf}$, and $ k_l+k_{rbf} + k_l*k_{rbf}$; and 3 kernels that encapsulated the noise in responses: $k_n$, $ k_n + k_l*k_n$, and $ k_n +k_{rbf}*k_n$.

Kernels were evaluated through 5-fold cross validation within each participant visit. The performance was averaged over all participant visits to evaluate each kernel. We select kernels based on their negative log marginal likelihood for the observed data. This value reflects the goodness-of-fit and additionally accounts for model complexity, but is only applicable to Bayesian models. We additionally examine other non-Bayesian models to compare accuracies and negative log likelihoods.

We found that modeling the interaction data with Gaussian processes reached similar levels of accuracy and negative log likelihoods as other machine learning models. For completeness, we also evaluated the use of other classifiers and regressors for calculating our metric for nonuse, and found them to perform similarly to Gaussian processes, as reported in the supplementary materials.

\subsection*{BARTR as a Metric of Nonuse}
To evaluate BARTR as a metric for neurorehabilitation, we evaluated the three criteria of effective metrics: validity, test-retest reliability, and ease of use.

\subsubsection*{Validity}
\begin{figure}[ht!]
  \centering
\includegraphics[width=\linewidth]{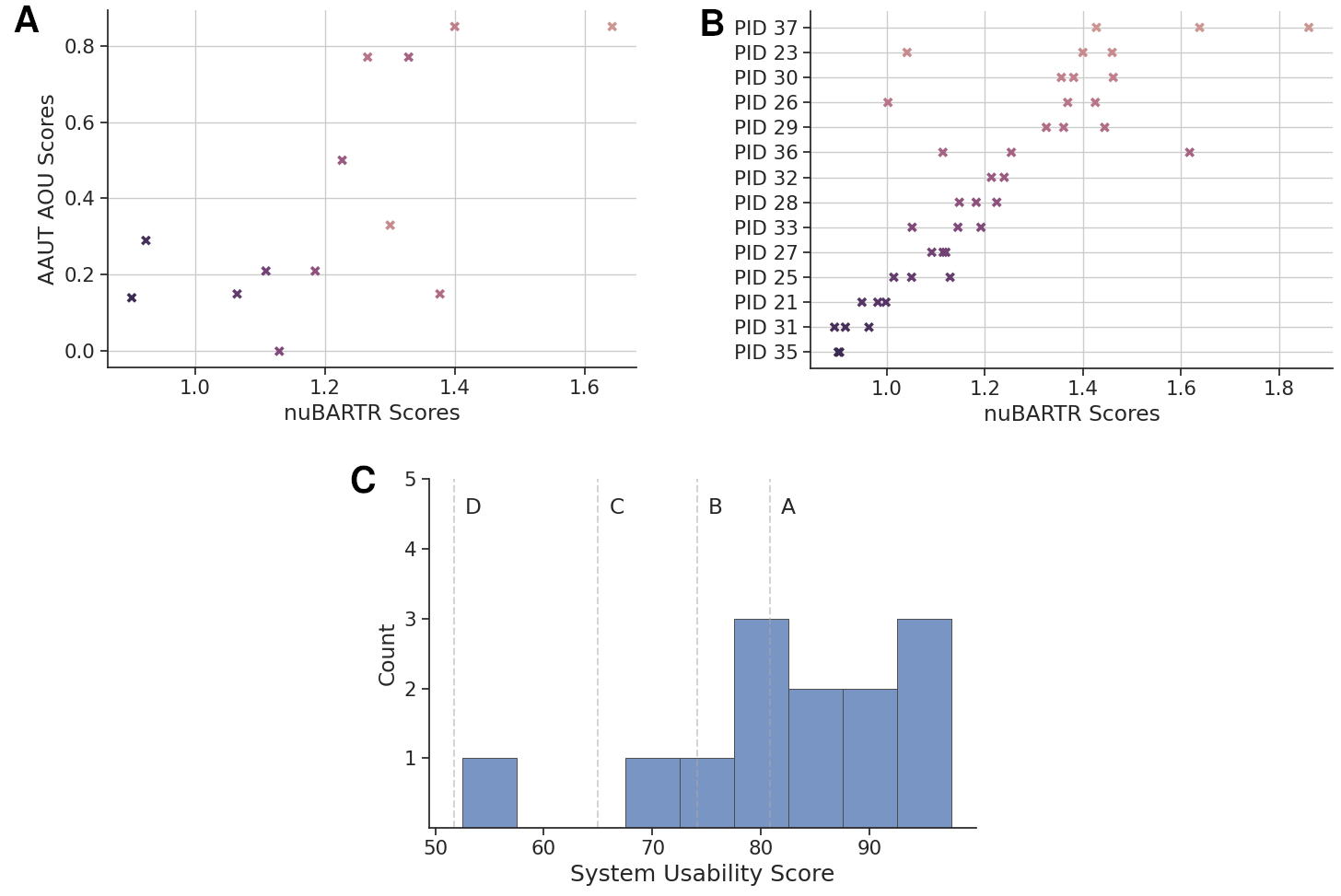}
      
    
    
  \caption{\textbf{Evaluations of the proposed metric.} We demonstrate the Bimanual Arm Reaching Task with a Robot (BARTR) metrics validity through its correlation with clinical measurements of nonuse through a non-parametric Spearman correlation, $r(13)=.693$, $p=.016$ (\textbf{A}). We demonstrate reliability with the absolute agreement of BARTR scores across three sessions through the intraclass correlation coefficient, $ICC(1,k)=.908$, $p<.001$ (\textbf{B}). We demonstrate its ease of use through usability ratings of the system, showing that the average rating is above 72.6 through a non-parametric Wilocoxon signed-rank test, $Z=16.0$, $p=.040$ (\textbf{C}).}
   \label{fig:evaluations}

\end{figure}

\nathan{We evaluated the validity of BARTR by comparing the quantification of nonuse produced by the system with the values of nonuse collected from post-stroke participants using the AAUT, the clinical standard for assessing nonuse.} Participants had a wide range of nonuse, with AAUT AOU values ranging from .00 to .85 and nuBARTR scores ranging from .849 to 1.71. \nathan{We determined the validity of nuBARTR with the non-parametric Spearman correlation between AAUT AOU and the averaged value of nuBARTR across the three sessions.} Figure \ref{fig:evaluations} shows that the calculated nonuse from BARTR is correlated with the clinical AAUT AOU metric of nonuse (r(13)=.693, $p=.016$). 

We also examined the correlation with the individual subscales of the AAUT with the non-parametric Spearman correlation. The cBARTR shows a high correlation with the cAAUT (r(13)=.773, $p=.002$) and the sBARTR shows a correlation with sAAUT (r(13)=.769, $p=.002$).

\subsubsection*{Test-Retest Reliability}
We examined the absolute agreement (ICC) of the three BARTR sessions to assess test-retest reliability. Absolute agreement of the BARTR metric is the recommended test of reliability in the medical field \cite{koo2016guideline}. We found that between sessions there was very high reliability of nuBARTR scores, ICC(1,k)=.908, $p<.001$. A visualization of nuBARTR scores by participant is shown in Figure \ref{fig:evaluations}. 

We noted correlations between all pairs of sessions via a Pearson correlation. The first and second session are significantly correlated (r(14)=.662, $p=.010$), the second and third sessions are significantly correlated (r(12)=.948, $p<.001$), and the first and third sessions are correlated (r(12)=.686, $p=.012$). We examined scores across all three sessions, and note that the BARTR interaction showed increased reliability after the first session, supporting repeated evaluations using this method to evaluate participants' nonuse over time. 

\subsubsection*{Ease of Use}
To evaluate ease of use, we applied the standard, commonly used System Usability Scale (SUS) \cite{brooke1996sus,bangor2008empirical,lewis2018system}.  The SUS is scored out of 100 and calculated from 10 items. SUS meta-analyses provide full distributions of SUS scores across 446 extant systems, and recommend evaluating systems based on percentiles of systems examined in the meta-analysis \cite{lewis2018system}. For example, a mean SUS score of 72.6 represents a system that is in the top 65\% of all systems evaluated in the meta-analysis, and the meta-analysis provides a rating system for understanding these percentiles. \nathan{A score of 78.9 or higher is in the `A' range, a score of 72.6 to 78.8 is in the `B' range, a score of 62.7 to 72.5 is in the `C' range, and a score of 51.7-62.6 is in the `D' range.} The middle values of these ranges are denoted by the dashed lines in Figure \ref{fig:evaluations}.

We administered the SUS to all participants that enrolled in the study. For determining usability, we examined the SUS scores of only the post-stroke group. The average rating of scores was 8.93 $\pm$ 11.67, placing the mean usability of the BARTR apparatus in the 80th percentile of systems included in the SUS meta-analysis. Due to the high variance in participants' scores, we determined that the score is significantly greater than 72.6, which corresponds to an above-average user-experience \cite{lewis2018system}. We found from a non-parametric Wilcoxon signed rank test that participants rated our system significantly above the 72.6 threshold (Z=16.0, p=.040). Based on this result, the system is easy to use, and readily satisfies the ease of use criterion. The distribution of SUS scores across all participants is shown in Figure \ref{fig:evaluations}.

\subsection*{Qualitative Results}
\begin{figure}[t!]
  \centering

      \includegraphics[width=.65\linewidth]{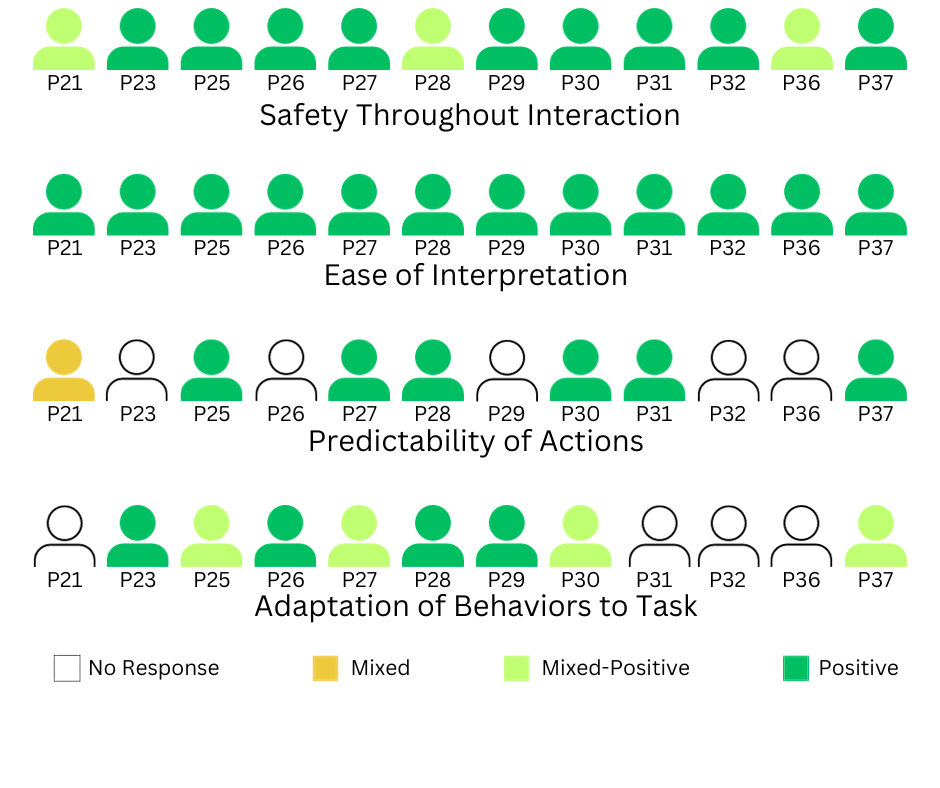}
      \vspace{-1cm}
      \caption{\textbf{Qualitative responses from participants.} We show overall perceptions of each of the four factors of trust \cite{kellmeyer2018social} that each participant mentioned.} \label{fig:qualitative_results}

\end{figure}

The qualitative analysis we performed considered the semi-structured interviews from 12 post-stroke participants who completed all three study sessions. The full set of interview questions is provided in the supplementary materials. The interviews were conducted following the third session of the BARTR, and lasted for an average of 14 minutes (minimum: 4 minutes, maximum: 44 minutes). The questions were structured around the four themes that prior work identified as important for interaction with rehabilitation systems~\cite{kellmeyer2018social}: safety throughout interaction, ease of interpretation, predictability of actions, and adaptation of behaviors to task.  We show an overview of the participants' responses to these four themes in Figure \ref{fig:qualitative_results}. Positive responses described the system as being unequivocally helpful within the theme, mixed-positive responses described the system as helpful but provided room for improvement, and mixed responses were unsure if whether the system was helpful with respect to the theme. No participants found the system unhelpful.  We also report the participants' suggestions for improvement and future tasks.

\subsubsection*{Safety throughout Interaction}
All participants ($n=12$) found the interaction to be safe. In addition to the safety precaution we took of moving the arm slowly, participants also reported feeling safe because they ``figured [the experimenter] knew what [they] were doing'' (P29) and that ``it felt pretty safe because I had this shoulder harness on'' (P27). 

Some participants ($n=3$) identified that they worried about the robot arm when it came close to the home position, but reported that this did not influence how safe they felt throughout the interaction. One participant viewed the perceived risk as beneficial to them because ``it was good to have my brain react to having it come close'' (P36).

\subsubsection*{Ease of Interpretation}
All participants ($n=12$) also found the robot easy to use. Most participants ($n=9$) specified that they felt this way because the interaction itself was \textit{easy to learn}. Participants found that they ``got used to the robot after the first command it gave'' (P37) and that the interaction was ``a normal everyday task, so it wasn't hard to learn'' (P27). Participants also found the task easy to learn because ``there wasn't anything ... that you have to put on'' (P23). Two participants (P29, P36) mentioned that they had done several other studies using other devices, and that this interaction was easy because ``it was all right in front of me and the instructions were clear'' (P29). 

Eight participants also directly described the socially assistive robotics's voice as easily understandable. One participant (P29) noted that they ``liked the mouth moving, it helped to understand the speech'' of the SAR providing instructions. In addition to understanding the words, another participant (P27) also found ``the voice was comforting and the instructions were very clear''. Participants found the instruction from the SAR valuable toward understanding the task as well as socially motivating.

\subsubsection*{Predictability of Actions}
Seven participants directly commented on the predictability of the interaction. The comments addressed both the physical predictability of the task, and the social predictability of the SAR. Participants found the task predictable because it was repetitive and simple. A participant (P37) found this to be particularly important because ``with stroke you're also going through a psychological situation, and with [this task], you don't have to grapple with anything. This way is straight-forward''.

Participants had a variety of interpretations of the social component of the interaction. Because we used randomness in the SAR's movements and feedback in order to make it appear more natural and life-like (as is standard in human-robot interaction work \cite{dennler2021personalizing,abubshait2020repetitive,de2015makes}), some participants viewed the unpredictability as a benefit. One participant (P27) referred to the unpredictable social behaviors as ``natural'' and thought the SAR ``doesn't feel like technology''; another participant (P25) became engaged in ``trying to find a pattern in the robot's eyes''. Another participant (P21) had a more neutral reaction to the randomness, and said ``the fluctuations in cuing, I don't know if that was a hindrance or a help''. One participant (P37) greatly appreciated the fact that the exercise was lead by a robot, because the overall social interaction was predictable and the robot was not getting tired, and stated ``with the SAR it is like no judgement...there is no feeling of changing in the delivery...if a person had to repeat `go, go, go', sometimes they might get tired, and when you're doing the exercise you can see that''.

\subsubsection*{Adaptation of Behaviors to Task}
Eight participants commented on how the system could adapt to them specifically throughout the task. The participants were also concerned with either the task or the social component of the task. For the task, six participants identified that the robot could adapt more to different levels of task difficulty. With the goal of developing a standardized test, the robot sampled points randomly in the interaction, but participants asked if the arm ``could go all the way up or all the way back...it would be nice if I could extend my whole arm'' (P25), while at the same time recognizing that ``if you had more damage in your arm it would be harder to do'' (P28). Four participants also described the timing of the robot placing points. Three of them found the speed appropriate; one of these participants (P37) described it as ``when the arm was moving it was moving at the right speed''. One participant (P21) thought that the arm ``could move faster or something...it was very methodical where it went''.

With regard to the SAR's verbal communication, four participants described the feedback that the SAR gave as evidence of it adapting to their good performance. P26 also specified that the SAR's progress updates were helpful because they ``gave you an idea of where you stand at the time''. However, one participant (P30) wished that the SAR would be ``more responsive'' to the specifics of their performance, for example through commenting on how fast their reach was.

\subsubsection*{Suggestions for Improvements and Future Tasks}
Participants also provided feedback on how the system could be improved or adapted to other forms of exercises for evaluating arm nonuse. The suggestions for improvement largely addressed how the system could be more personalized to individual tastes. Participants discussed how visual components could be adapted, for example how the SAR's exterior could ``change to USC colors, which would work better... I have some stickers I could put on the robot'' (P37), or how the button could ``turn green when you press it'' (P23). Other participants described how the SAR's audio could be personalized by ``choosing music to play, just to make it more pleasant'' (P25). Participants also suggested gestures for the robot to perform, such as ``when you make a mistake, you could have the robot hold its arms up and point to the button'' (P37).

Despite these suggestions, participants (n=10) described the system as being effective and helpful. Several (n=4) explicitly stated that they thought about the interaction outside of the experiment. One participant reported that when they were ``trying to open a cabinet, I had a flashback to this button pressing when I was thinking about how to orient my hand to open the cabinet'' (P37). Participants found the ``fact that it is 3D is effective'' (P37), and suggested several other three-dimesional interactions that would be useful.

The most popular task that the participants described as being useful was ``3D tasks that involved more finger dexterity'' (n=7). Participants described how the interaction could ``integrate a little ball... because once you put it in your hand your fingers start working'' (P36), or how the robot arm could ``hold a pocket or something and have people put pennies over here or over there'' (P37). One participant also described how they would like to control the robot to practice finger dexterity by using ``a glove or something to control the robots, so you simulate grabbing and the robot moves with the glove'' (P23). 

The second type of task that multiple participants suggested was gross motor tasks (n=4). For example, two participants suggested using the robot arm to passively move their more stroke-affected side by ``grabbing what the robot is holding and have it drag my arm around'' (P25). Two other participants suggested actively pushing against the arm as a form of strength training. One participant suggested ``you could add on pressure sensing...I am interested in seeing the pressure and strength of both sides'' (P27).

%
%

\section*{Discussion}
This work introduced robotics as an enabling methodology for the evaluation of difficult-to-evaluate yet clinically substantial constructs such as arm nonuse post-stroke. We demonstrated that robots can provide a way to objectively and reliably assess motor behaviors that are meaningful for neurorehabilitation. The following discussion highlights the efficacy of using a SAR and a robot arm together for rehabilitation and evaluation to complement the work of neurorehabilitation clinicians. The quantitative and qualitative results and insights contributed by this work, including the interviews with post-stroke participants following their BARTR sessions, inform the design of future systems for effective rehabilitation and assessment of patients' rehabilitation progress.  

\subsection*{Robots as Tools for Evaluation}


A key benefit of using a robotic system for administering rehabilitation assessments is in enabling highly controlled, repeatable, and precise measurements. Robot arms/end-effectors can administer tests with exact instruction, intonation, pacing, and placement of reaching targets, as well as randomize variables that may affect outcomes, allowing these variables to be assessed covertly, an important aspects of valid assessment methods. Additionally, SARs can provide instruction and motivation for sustained effort of long-term rehabilitation exercises. Users do not need to wear any sensors, allowing for unencumbered, natural behavior more representative of ADLs, a key consideration that allows the BARTR to achieve high reliability and usability compared to other interactions that rely on wearable sensors to collect data \cite{boukhennoufa2022wearable}, as worn sensors can affect behavior by being encumbering or uncomfortable \cite{yin2021wearable}.

In the context of other metrics, using robotics provides the benefit of precise spatial information, enabling quantification of areas of difficulty for arm use. Precise quantification of those regions can be tedious or difficult for clinicians to obtain, yet the data can support the development of personalized therapy regimens. Once the regions are quantified, they can be used to adapt the system's behavior to select targets for the user that are at the appropriate level of challenge, with sufficient variety as well.  Another alternative is to use environmental or ambient sensors to track human movement, but such sensors can present privacy concerns.

With the increased richness of the resulting data from robotic systems administering physical assessments, more nuanced information can be communicated to rehabilitation therapists and clinicians. For example, machine learning techniques can be applied to summarize a patient's data for easy visualization by clinicians, who can then indicate where in the workspace the patient may need to focus most. This can be connected with data visualization techniques to effectively communicate particular measurements of interest and enable more personalized care.

Past research has also shown that end-effector robots can be functionally effective in rehabilitation \cite{lee2020comparisons}. This work further demonstrates that such robots have the potential to be used for assessments that are otherwise difficult to obtain due to lack of reliability or ease of use. Our metric of nonuse is not specific to a particular robot embodiment, and can be applied to any robot that can reach points within the 3D space around the patient in a cylindrical region 30 cm in radius and 40 cm in height (as illustrated in Figure \ref{fig:workspace}). Such robots can be used for other forms of assessment \cite{balasubramanian2012robotic}, adaptive exercise practice that require a model of the difficulty of reaching different points in the participant's workspace \cite{blank2014current,mounis2019assist}, and other gamified tasks to practice reaching \cite{eizicovits2018robotic}. In addition to rehabilitation exercise, other assistive tasks throughout the rehabilitation process can be completed by end-effector robots, such as dressing\cite{jevtic2018personalized}, hair combing\cite{dennler2021design}, shaving\cite{hawkins2014assistive}, etc. Due to their multiple possible uses and their portability, end-effector robots have the potential to facilitate the in-home rehabilitation process. Analogously, SAR systems have been shown to be effective in increasing user motivation for a wide variety of tasks, from physical exercise \cite{fasola2013socially} to cognitive and social skill learning \cite{clabaugh2019long}, an in early work on SAR for supporting stroke rehabilitation exercises \cite{mataric2007socially}. This paper demonstrates a synergistic combination of both a robot arm and a SAR in a rehabilitation context.

\subsection*{Considerations for Using Robots for In-Home Assessments}

Incorporating robots into home environments is a broad and highly active area of robotics research. In-home rehabilitation, also called in-place rehabilitation, is a frontier that presents a unique set of challenges. Naturally, safety is a key concern. In this work, a robotics specialist was present for all sessions to monitor any potential system failures. Over 8,000 trials were conducted as part of this work, and there was only one case of the robot moving outside of the workspace.  The error was quickly corrected, but such expert intervention would not be readily available for in-home systems. Although all our study participants reported not feeling scared or anxious by the robots' movements, future unsupervised systems must be developed with considerations for all failure cases, no matter how improbable.

Another concern is the privacy of the collected user data. Some data used in this work were personally identifiable--for example, the robot used the participants' names when addressing them, which was appreciated by participants who noted that it added to the interaction (P25). Although identifiable data are valuable for engaging and personalizing interactions, they also represent a data security risk.

Another barrier to bringing robots to users' homes is cost. Design guidelines for robotic systems for rehabilitation therapy indicate that to be useful in home settings, such systems must cost less than US\$10,000 \cite{van2018assistive}. Hospital loans and insurance coverage are necessary to subsidize patient costs. Low-cost (under US\$1,000) SAR systems have already been developed \cite{suguitan2019blossom}, and more cost-effective end effector robots are also being developed \cite{robotics_2023}. In our work, post-stroke participants reported that they would miss the robot after their third session. If the system is used for a longer period of time in the home, a stronger bond may develop, and thus future work should evaluate how removal of the system may affect users \cite{taylor2013leaving}.

\subsection*{Limitations and Future Work}

Although the AAUT is still used for clinical evaluation of nonuse, it may have become outdated. It includes tasks such as inserting a Polaroid into a photo album and opening a physical newspaper; such tasks are commonly accomplished digitally and therefore may compromise the covert nature of the test. Although no participants reported being aware that they were being observed before the nature of the test was revealed, outdated/atypical tasks may also influence arm use due to task unfamiliarity within the context of activities of daily living. The question of aptness of the AAUT provides a further impetus to develop additional assessments of nonuse, such as the BARTR.

\nathan{Although we show that the BARTR possesses the criteria for a good neurorehabilitation metric, we note that there are two important considerations to make for its use in clinical contexts. First, we observed much higher reliability in the second and third BARTR sessions, and variability in some of the participants between sessions. We recommend one abbreviated session to habituate the participants to the BARTR assessment for increased reliability. Second, the BARTR is focused on reaching motions and does not include finger/hand manipulation tasks.} Therefore, a limitation of the presented study is that the button task was relatively simple compared to the fine manipulation tasks used to evaluate nonuse in the AAUT, such as removing business cards from a box and placing a picture in a scrapbook. Such tasks may not be correctly evaluated through our current system, however future extensions of BARTR test can include fine manipulation tasks. Our qualitative results show participants suggesting the inclusion of additional held objects as part of the BARTR trials, in order to enable the assessment of grasping and stabilization motions in addition to reaching motions \cite{varghese2020probability}. The participants suggested tasks that involved the manipulation of pennies and golf balls; the BARTR could include a screwdriver-like instrument to press the button, analogous to the cylindrical grasp item of the FM-UE. Such additional tools can also be equipped with sensors to evaluate grasp force, which cannot be evaluated visually \cite{choi2011feasibility}.


Our results are limited to only right-hand dominant participants as we did not recruit any left-handed post-stroke participants in order to reduce variance in our tested population, since handedness affects baseline hand choice. Previous work has shown that the side of the stroke lesion is more important in determining limb selection over pre-morbid handedness \cite{kim2022effort}. As more data are collected from BARTR sessions, stronger results can be drawn about pre-morbid handedness in this specific task. However, due to well-known difficulties in recruiting pre-morbidly left-hand dominant stroke survivors, data scarcity may be a barrier for evaluating all participants' nonuse. A potential direction may be to evaluate if techniques from domain adaptation \cite{wilson2020survey} may be used to leverage data collected from right-handed participants to perform more accurate assessments of future left-handed users.

Our qualitative results are influenced by the particular identity of the analyst. One author (ND) conducted the interviews and analyzed the interview data. In the spirit of reflexivity \cite{berger2015now}, it is important to understand that the analysis represents the views of a non-disabled computer scientist that designed the interaction and values the co-construction of knowledge with participants. ND has experience both as a facilitator and a participant of participatory design in several contexts and views the feedback of the users of systems as a requirement in the design process.

The most important direction for future work is to deploy and evaluate this methodology for longer periods of time as a tool for clinicians to assess and develop progress throughout the patients' rehabilitation process. Several opportunities for further research would arise from such deployments. Researchers could examine how information is communicated to neurorehabilitationists, how information from the BARTR test can improve functional outcomes or other behavioral metrics in users, how SAR personality and communication styles may be adapted to participants over time, how BARTR tests could use data from previous sessions with the same participant to be more efficient and engaging, and how assessment can be combined with regular rehabilitation exercises. Overall, the combination of social and functional components offers a unique opportunity for more personalized, engaging, and effective human-robot interaction for neurorehabilitation.

%
%

\section*{Materials and Methods}
\subsection*{Participants}\label{RecruitmentSection}
Participants who had experienced a hemispheric stroke at least 6 months before enrollment were recruited for this study. Participants were screened for eligibility prior to the interaction and were deemed eligible if they satisfied the following criteria: 18 years of age or older,able to reach at least 30cm anterior to the mid-line of the trunk and at least 30 centimeters high without pain or assistance, normal or corrected to normal hearing and vision, had proficiency in understanding English, were right-hand dominant pre-stroke, and had a score of greater than 25 out of 66 on the Fugl-Meyer Upper Extremity Motor Assessment \cite{fugl1975method}.

The Fugl-Meyer assessment was administered to all participants by a board-certified physical therapist specializing in neurorehabilitation who had more than two years of experience. We also administered the Mini-Mental State Exam (MMSE) \cite{folstein1975mini} to ensure that the participants could give consent. \nathan{If a participant scored lower than 25 of 30 on the MMSE, a caregiver was required to be present as a witness for consent.}

We additionally recruited neurotypical participants of similar ages to establish normative use of the robot system. These participants provided a baseline for zero nonuse, because neurotypical participants favor their dominant hand in bimanual tasks \cite{han2013quantifying}. The purpose of the neurotypical adult group was to establish a normative value for handedness bias as well as the time it takes to reach different points in the task workspace.

All participants reviewed and signed a consent form prior to the experiment. Participants with chronic stroke performed up to three sessions using the BARTR testing apparatus, scheduled at least 4 days apart. Neurotypical adults performed one session with the BARTR testing apparatus. Participants were paid 50 USD for each hour-long session they completed. All study protocols and consent forms were approved by the University of Southern California Institutional Review Board under \#UP-22-00461.

\subsection*{Actual Amount of Use Test}
Prior to the first session of BARTR, the experimenter (ND) administered the Actual Amount of Use Test (AAUT) to the chronic stroke survivors, and the research physical therapist (AC) rated performance from the offline video data. The AAUT is a covert assessment of spontaneous arm use for 14 tasks that regularly occur in daily life, such as pulling out a chair from a table prior to sitting in it and flipping through the pages of a book. First the tasks were completed covertly (spontaneous AAUT$_s$), \nathan{so the participant did not know that they were being video recorded and tested.} Then, the experimenter revealed that arm use was being observed, and participants completed the 14 tasks again while being encouraged to use their stroke-affected arm as much as possible (constrained AAUT$_c$).

The research physical therapist rated both the AAUT-Amount (binary yes/no) if the participant attempted to use their stroke-affected arm (AAUT AOU score) for that task and AAUT-Quality of Movement or QOM (on an ordinal scale of 0 to 5) if they used the paretic arm in the task. In the context of this study, we considered the AAUT AOU score as the metric of interest for three reasons. First, the AOU and QOM scales have been found to be redundant \cite{uswatte2005implications}. Second, in our sample, the AOU achieved higher values of internal consistency for each participant ($\alpha=.87$) than the QOM scale ($\alpha=.72$). Finally, the post-stroke participants exhibited a high coverage of the values of the AOU scale. The final nonuse score for each participant was calculated as the average of the differences between the constrained AAUT AOU score and the spontaneous AAUT AOU score over all tasks, resulting in a scalar value between 0 and 1.

\subsection*{Testing Apparatus}

The BARTR apparatus, designed to test arm nonuse, consists of a robot arm and a socially assistive robot (SAR). The robot arm was the Kinova JACO2 assistive arm \cite{kinova} selected because it has already been used in assistive domains, it is lightweight, and it safely interacts with and around people. The arm has the same affordances as end-effector robots typically used for other rehabilitative interactions that have been shown to be effective in the rehabilitation context \cite{lee2020comparisons}. The SAR was the Lux AI QTRobot \cite{luxai} that consists of a screen face on a 2 degree-of-freedom head and two 3 degree-of-freedom arms that can gesture. This SAR platform has already been validated in our past work with children with arm weakness due to cerebral palsy \cite{dennler2021personalizing}, as well as in other human-robot interaction contexts \cite{spitale2022socially}. The SAR provided the participant with verbal instructions at the start of the BARTR session, and with positive feedback on a random schedule, similarly to previous SAR use in other rehabilitation contexts \cite{swift2015effects,dennler2021personalizing,feingold2021robot}. 

In addition to the two robots, we developed two low-cost devices for the BARTR apparatus: the target object and the home position. Both devices are 3D-printed, have self-contained power supplies and processors, and communicate wirelessly with the BARTR apparatus using low-level UDP protocols. 

The target device, held by the robot arm, consisted of a 3D-printed housing with a single button. It received commands to turn on a light and start a timer to begin each reaching trial, and logged the time taken by the participant to reach for and press the button to turn off the light.

The home position was the location that participants returned to between reaching trials, implemented as a 3D-printed block with two shallow holes 2cm in diameter, 5 cm apart, with capacitive touch sensors inside. Participants placed their left pointer finger in the left hole, and their right pointer finger in the right hole. The device communicated at 20Hz, reporting the locations that were being actively touched by the participant.

\begin{figure}[t!]
  \centering
    \includegraphics[width=\linewidth]{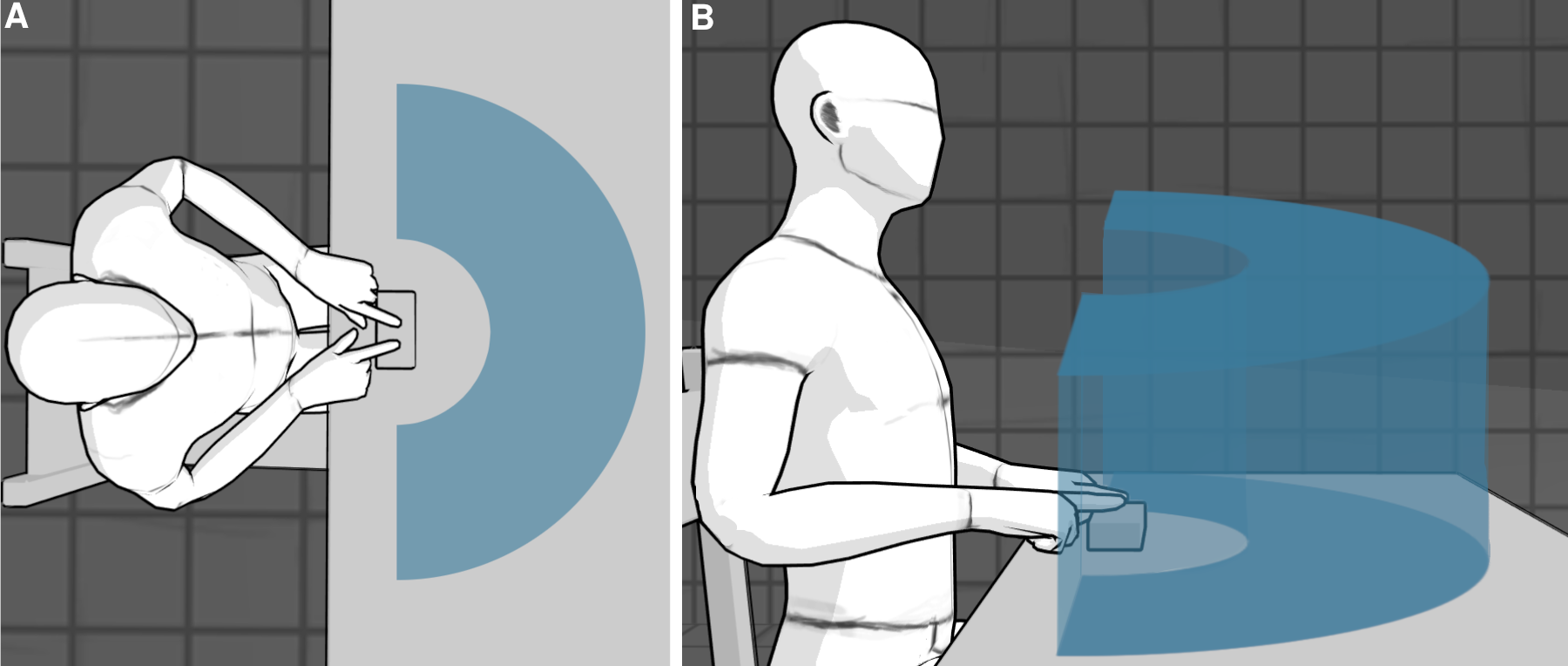}
    
  \caption{\textbf{Visualization of the participant's workspace.} Viewed above, the workspace tested extends radially from the home position from a distance of 10cm to 30cm (\textbf{A}). Viewed from the side, the workspace extends upward 40cm (\textbf{B}).}
   \label{fig:workspace}

\end{figure}

\subsection*{Bilateral Arm Reaching Test with a Robot}\label{BARTR}

Participants were seated at a table with the home position aligned with the center of their chest, as shown in Figure \ref{fig:workspace}. They were instructed to maintain approximately 90 degree angles of their elbows when their index fingers were resting at the home position. To limit upper-body compensatory movement, participants wore a shoulder harness attached to the chair \cite{cai2019detecting}. Participants were instructed to verbally cue the experimenter when they were ready to begin each section of BARTR. Following previous work, the two experiment phases were the spontaneous BARTR phase (sBARTR), where the participants were instructed to use either arm to reach the target, and the constrained BARTR phase (cBARTR), where the participants were instructed to use their more-affected arm to reach the target \cite{han2013quantifying}.

\nathan{For both phases of BARTR, the robot arm placed the reaching target at a different location in 3D space in front of the participant.} \nathan{The participant was instructed to reach the target as quickly and accurately as possible when prompted by the SAR}. Each reaching trial began with the robot arm moving to one of the randomly sampled locations. When the robot arm arrived at the location and the participant was in the home position, the light on the target device turned on and the SAR cued the participant to reach to the target after a random interval between 0 and 2 seconds, to prevent the participant from anticipating movement to the target. After the audiovisual cue, the participant was given 3.1 seconds to reach to the target. \nathan{When the participant pressed the button, the light turned off. If the participant did not reach the target in 3.1 seconds, the light turned off after the 3.1 seconds had elapsed.} This time period was selected to make the maximum time of each experiment phase approximately 20 minutes in duration, given the variability in travel time between points for the robot arm. This period was sufficient for all neurotypical participants to reach all of the target placements.

In total, 100 locations were tested for each of sBARTR and cBARTR. The locations were evenly spaced in the 3D workspace volume in front of the participant defined by the region that was 10cm to 30cm from the center of the home position, forming a semi-circle that extended in front of the participant in their transverse plane, and heights that ranged from 0cm to 40cm above the table, as shown in Figure \ref{fig:workspace}. These points were selected randomly without replacement--namely, each point was selected exactly one time; participants reached for all 100 targets one time per session. Participants attempted up to 100 reaching trials for each section of BARTR, for a total of 200 reaches.

\subsection*{Calculation of the BARTR Metric}\label{calculation}

We used the data collected through the BARTR interaction to estimate a user's workspace. Following previous work, nonuse was modeled as the subtraction of two components: the constrained component and the spontaneous component \cite{han2013quantifying}. The constrained component of the workspace $W$ is defined for every point $x \in W$ for a particular participant $p$ as:
\begin{equation}
    cBARTR_p(x) = p_{p}(success | X=x, S=s_p)
\end{equation}
where $p_p(\cdot)$ denotes the function that returns the probability of the poststroke participants selecting each side in the spontaneous condition. The side of the participant that was affected by stroke is denoted as $s_p$ and is in the set of values $\{'left', 'right'\}$. This quantity represents the total area that the participant is expected to be able to reach within the time limit--3.1 seconds, based on the times from the neurotypical group. 

The spontaneous component of the workspace is defined over all points $x \in W$ as
\begin{equation}
    sBARTR_p(x) = p_{p}(S=s_p | X=x)*p_{n}(S=s_p | X=x)*E[t^{s_p}_{n}(x) - t^{s_p}_{p}(x)|X=x]
\end{equation}
where $p_p(\cdot)$ denotes the probability of the post-stroke participants selecting either side in the spontaneous condition and $p_n(\cdot)$ denotes the probability of the neurotypical group selecting either side in the spontaneous condition. $t^{s_p}_{p}(x)$ and $t^{s_p}_{n}(x)$ represents the movement time for the post-stroke and neurotypical participants, respectively, to reach the point $x$ in the workspace with the arm on the participant's more affected side, $s_p$. This quantity represents how close the participants' spontaneous arm use is to spontaneous neurotypical use. Higher usage of the participant's more-affected arm, and faster movements result in higher spontaneous scores.

The final calculation for nonuse is calculated as the difference of these functions summed over all of the points in the workspace:
\begin{equation}
    nuBARTR = \sum_{x \in W} cBARTR(x) - sBARTR(x)
\end{equation}
To obtain these values, we modeled the interaction metrics--time to reach points and arm choice--as Gaussian Processes for the normative participants and for each post-stroke participant. We summed over 10,000 samples \nathan{from a uniform distribution over the} workspace to accurately estimate the difference of these two functions.

\subsection*{User-Reported Data}
In addition to the data-driven evaluation of nonuse, we asked participants for their perceptions about using the system with two self-reported surveys: the System Usability Scale \cite{brooke1996sus,bangor2008empirical,lewis2018system} and a semi-structured interview \cite{kallio2016systematic}.

\subsubsection*{System Usability Scale}
The SUS \cite{brooke1996sus} is a 10-item scale that assesses the ease of use of a technological system. Each item is rated on a 5-point Likert scale that ranges from "Strongly Agree" to "Strongly Disagree". Five of the items are positively worded, where higher ratings indicate a highly usable system, and five items are negatively worded, where lower ratings indicate a more usable system. This scale was selected for its high reliability, validity, and broad applicability to technological systems. \nathan{Participants completed the SUS following their second session with the system.}

\subsubsection*{Semi-Structured Interviews}
The combination of a SAR and a robot arm has the potential to support participants throughout their rehabilitation process at home. Due to the socially interactive nature of the BARTR session, rehabilitation effectiveness depends on how much users can trust the robots to help them. Trust in rehabilitation robotics has four key facets: safety throughout the interaction, predictability of actions, ease of interpretation, and adaptation of behaviors to the task \cite{kellmeyer2018social}. We developed questions to identify ways that the system we developed could support participants in other exercises and how the system they interacted with achieved or did not achieve the above four key aspects of trust. \nathan{Participants qualitatively reflected on their experience of the system and discussed improvements for future systems following the third session.}

\nathan{One of the authors conducted and analyzed all of the interviews. We used an iterative, four-phase deductive qualitative analysis approach to analyze the interview data \cite{elo2008qualitative}. The first phase consisted of the transcription and open coding of interview data, assigning specific meaning to the phrases participants spoke. The second phase grouped similar codes into sub-themes. The third phase categorized the sub-themes according to the themes found in previous research in rehabilitation with SARs \cite{kellmeyer2018social}. Some codes did not fit into the theorized categories, and thus new emergent categories were developed. The fourth phase iterated on the second and third phases with developing categories on a subset of 6 interviews that were expanded to include all the interviews until the final categorization reached theoretical saturation, similar to the method used by Ando et al. \cite{ando2014achieving}.}

\bibliography{scibib}

\bibliographystyle{Science}

\textbf{Acknowledgments:} We thank all participants and pilot testers that participated in our study for their time. \textbf{Funding:} This project is supported by a National Science Foundation Graduate Research Fellowship (\#DGE-1842487). C.W. was supported in part from the NIH R41 Small Business Technology Transfer funds (\#HD104296) and NIH R01 (\#HD059783). S.N. is supported by an Agilent Early Career Professor Award. M.M. is supported in part by NRI: FND: Communicate, Share, Adapt: A Mixed Reality Framework (\#GR1014084).
\textbf{Author contributions:} N.D. designed and performed all experiments and wrote the manuscript; A.C. administered Fugl-Meyer assessment and scored Actual Amount of Use Test and provided feedback on experimental design; E.D. and C.C. developed code for analyzing results; C.W. provided the lab space, participants, and collaboratively developed the experimental protocol and paper editing; S.N. helped with analysis plans and paper editing; and M.M. helped design experiments and edited the paper.
\textbf{Competing interests:} C.W. is a member of the data safety and monitoring board for Enspire DBS Therapy, Inc (DBS is Deep Brain Stimulation), and Brain Q (Syntactx). She receives an honorarium for her services. She is a member of the external advisory board for MicroTransponder, Inc., MedRhythm, Inc., and Axem Neurotechnology, Inc. She receives payment for her consulting. She is Editor of the 6th edition of Motor Control and Learning, published by Human Kinetics, Inc and receives royalty payments. She is an Editor for the 2nd Edition of Stroke Recovery and Rehabilitation, published by DemosMedical Publishers and receives royalty payments. 
\textbf{Data and materials availability:} All data needed to evaluate the conclusions in the paper are present in the paper or the Supplementary Materials. The data for this study will be made available upon request in accordance with our IRB protocol.

\end{document}



\baselineskip24pt


\maketitle 

\newpage

\section*{BARTR Interaction Validity Check}\label{validity_check}
\begin{figure}[h!]
  \centering
    \includegraphics[width=\linewidth]{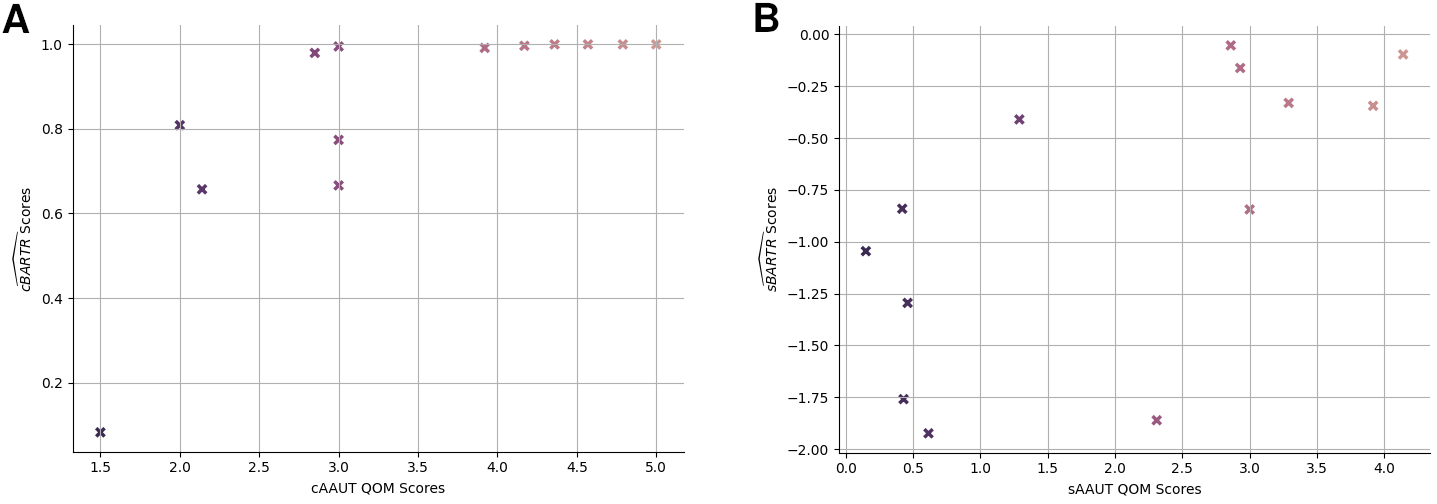}
  \caption{\nathan{\textbf{Correlations of Bimanual Arm Reaching Task with a Robot (BARTR) metrics from raw interaction data and corresponding Actual Amount of Use (AAUT) test metrics.} Estimates of constrained BARTR metrics show significant correlation with constrained AAUT scores through a non-parametric Spearman correlation, $r(13)=.896$, $p<.001$ (\textbf{A}). Spontaneous BARTR scores are also correlated with spontaneous AAUT scores through a non-parametric Spearman correlation, $r(13)=.610$, $p=.027$ (\textbf{B}).}}
   \label{fig:raw_correlations}

\end{figure}

We analyzed the raw data from the two phases of the (BARTR) session for correlation with the corresponding phases of the AAUT to verify that our results are not influenced by our choice to model the participants' interaction metrics across the space. The constrained phase of the AAUT is meant to capture functional capability of the affected arm. From the raw data, this is estimated as the number of successful button presses in the constrained condition of the BARTR:

\begin{equation}
    \widehat{cBARTR} = \frac{n_{success}}{n}
\end{equation}

where $n_{success}$ denotes the number of successful reaching trials in the constrained phase of the BARTR, and $n$ denotes the number of total reaching trials in the constrained phase of the BARTR. We found that $\widehat{cBARTR}$ was strongly correlated with the cAAUT via a non-parametric Spearman correlation r(13)=.896, $p<.001$, \nathan{illustrated in Figure \ref{fig:raw_correlations}}.

The spontaneous phase of the AAUT captures the automatic capability of arm use compared to the neurotypical behavior. Higher spontaneous scores occur when chronic stroke survivors use their affected side as often as the neurotypical population uses the same side. We estimated this value from the raw hand choice data from the spontaneous phase, and from the time to reach with the affected arm from both phases, as follows:

\begin{equation}
    \widehat{sBARTR} = min(\frac{n^{p}_{success,s_p}}{n^{n}_{s_p}}, 1) \cdot min(t^{n}_{s_p} - t^{p}_{s_p},0)
\end{equation}

where $n^{p}_{success,s_p}$ denotes the number of successful reaching trials performed with the more affected side of the chronic stroke participant, and $n^{n}_{s_p}$ represents the average number of times the neurotypical group used their hand on the same side. $t^{p}_{s_p}$ denotes the average time across all reaching trials when the chronic stroke participant used the hand on their affected side to press the button, and $t^{n}_{s_p}$ represents the average time to complete the trial in the neurotypical group using the hand that corresponded with the stroke participant's more affected side. These values are constrained because the spontaneous score cannot be better than the normative use in the neurotypical group. We find that $\widehat{sBARTR}$ is significantly correlated with the sAAUT via a non-parametric Spearman correlation r(13)=.610, $p=.027$, \nathan{illustrated in Figure \ref{fig:raw_correlations}.}

\section*{Metric Results for Nonuse using Other Modeling Choices}\label{SVMResults}
We examine the sensitivity of our metric for nonuse by considering other methods for modeling the success probability and the side choice probability. Specifically, we examined the best performing models for each modeling problem as determined by accuracy in the manuscript. and also consider the best-performing non-Gaussian process approaches. The results are summarized in Table \ref{table:alt_results}.

\begin{table}[!ht]
    \centering
    \hspace*{-3em}
    \begin{threeparttable}
    \caption{Metric Evaluations for Other Modeling Choices} \label{table:alt_results}
    \begin{tabular}{ccc|cc}
    \hline
Choice Model & Success Model & Time Model & Correlation with AAUT AOU & ICC(1,k) \\
\hline
GP           & GP            & GP         & 0.693*              & 0.908*** \\
MLP          & RBF SVM       & GP         & 0.662*              & 0.889*** \\
RBF SVM      & RBF SVM       & GP         & 0.687**             & 0.848*** \\
MLP          & RBF SVM       & MLP        & 0.715**             & 0.881*** \\
RBF SVM      & RBF SVM       & MLP        & 0.701**             & 0.843***\\ 
\hline
\end{tabular}
    \begin{tablenotes}
      \small
      \item First row represents evaluation used for analysis, subsequent rows show different modeling choices for each model. * denotes $p <.05$, ** denotes $p < .01$, *** denotes $p<.001$. Abbreviations: ICC, intraclass correlation; MLP, multi-layer perceptron; RBF SVM, radial basis function support vector machine. 
    \end{tablenotes}
    \end{threeparttable}
\end{table}

We found that all modeling choices resulted in similar correlations with the AAUT AOU and similar test-retest reliability scores. Notably, the Gaussian Process approach showed the highest test-retest reliability score, likely because Gaussian Processes provide more conservative and probabilistically valid estimates of hand choice and success, which are particularly noisy in real-world scenarios. The other methods are likely to overfit to the specific interaction and provide more variable results across sections.

\newpage

\section*{Questions for User-Reported Surveys}\label{app:user_reported_measures}

\begin{table}[!ht]
    \centering
    \caption{System Usability Questions.}
    \begin{threeparttable}
        \begin{tabular}{|l|}
            \hline
            \textbf{1. }I think that I would like to use this system frequently.\\
            \textbf{2. }I found the system unnecessarily complex.\\
            \textbf{3. }I thought the system was easy to use.\\
            \textbf{4. }I think that I would need the support of a technical person to be able to use this system.\\
            \textbf{5. }I found the various functions in this system were well integrated.\\
            \textbf{6. }I thought there was too much inconsistency in this system.\\
            \textbf{7. }I would imagine that most people would learn to use this system very quickly.\\
            \textbf{8. }I found the system very cumbersome to use.\\
            \textbf{9. }I felt very confident using the system.\\
            \textbf{10. }I needed to learn a lot of things before I could get going with this system.\\
        \hline
        
        \end{tabular}
    \end{threeparttable}
\end{table}

\begin{table}[!ht]
    \centering
    \caption{Semi-structured Interview Questions.}
    \begin{threeparttable}
        \begin{tabular}{|l|}
            \hline
            \textit{\textbf{Safety throughout Interaction}}\\        
            \textbf{A.} Did interacting with system make you feel unsafe or put you on edge at all? Why or why not? \\
            \textbf{B.} Did your sense of safety change at all during the interaction?\\ 
            \\
            \textit{\textbf{Predictability of Actions}}\\
            \textbf{A.} How do you view the relationship between the robot arm and the socially assistive robot?\\
           \textbf{B.} What did you think of the robot arm placing targets for you to reach towards?\\
            \textbf{C.} Do you think this system could be helpful to practice different exercises over doing them alone?\\ 
            \\
            \textit{\textbf{Ease of Interpretation}}\\
            \textbf{A. }What did you think about robot providing instructions and feedback throughout the exercises?\\
            \textbf{B. }Do you have previous experience with new technologies? \\
            \textbf{C. }How easy or difficult was learning to use this in comparison with those other new technologies?\\
            \\
            \textit{\textbf{Adaptation of Behaviors to Task}} \\
            \textbf{A. }Did you notice any differences in the way the robots behaved between your visits?\\
            \textbf{B. }What other interactions would you like to have with this system?\\
        \hline
        
        \end{tabular}
    \end{threeparttable}
\end{table}

The full set of coded responses will be made available to researchers upon request.